\documentclass[11pt,a4paper]{article}
\usepackage[hyperref]{acl2020}
\usepackage{lipsum}
\usepackage{times}
\usepackage{latexsym}
\usepackage{graphicx} 

\usepackage{algorithm}
\usepackage[noend]{algpseudocode}

\usepackage{microtype}

\aclfinalcopy 

\newcommand\blfootnote[1]{%
  \begingroup
  \renewcommand\thefootnote{}\footnote{#1}%
  \addtocounter{footnote}{-1}%
  \endgroup
}
\title{Modelling Bahdanau Attention using Election methods aided by Q-Learning}
\author{Rakesh Bal* \\ Indian Institute of Technology\\ Kharagpur, India \\ \texttt{rakesh.bal@iitkgp.ac.in} \And
              Sayan Sinha* \\ Indian Institute of Technology\\ Kharagpur, India \\ \texttt{sayan.sinha@iitkgp.ac.in}
              }

\date{}

\begin{document}
\maketitle
\begin{abstract}
Neural Machine Translation has lately gained a lot of ``attention" with the advent of more and more sophisticated but drastically improved models. Attention mechanism has proved to be a boon in this direction by providing weights to the input words, making it easy for the decoder to identify words representing the present context. But by and by, as newer attention models with more complexity came into development, they involved large computation, making inference slow. In this paper, we have modelled the attention network using techniques resonating with social choice theory. Along with that, the attention mechanism, being a Markov Decision Process, has been represented by reinforcement learning techniques. Thus, we propose to use an election method ($k$-Borda), fine-tuned using Q-learning, as a replacement for attention networks. The inference time for this network is less than a standard Bahdanau translator, and the results of the translation are comparable. This not only experimentally verifies the claims stated above but also helped provide a faster inference.
\end{abstract}
\section{Introduction}
\blfootnote{* Equal contribution}
Attention mechanism has given rise to new and improved language models which are prominent in a wide range of downstream tasks such as machine translation, question-answering etc. However, with their improved performance comes the disadvantage of their massive size and increased complexity. This has hindered their adoption on a full-fledged basis in both academia and industry. In this work, we attempt to model one of the first attention networks using election methods and Q-learning to achieve lesser complexity with only a slight compromise in performance. We focus our experiments on Neural Machine Translation (NMT) using Bahdanau attention network to serve as a proof of concept. We also show that this reduced complexity results in a decrease in inference time.

\begin{figure}[t]
\centering
\includegraphics[width=0.4\columnwidth]{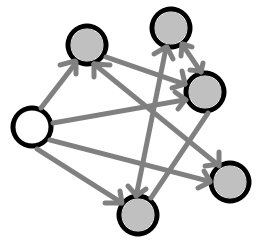} 
\caption{Example state space: The white and grey filled circles represent output and input words, respectively. Edges exist among words having a high cosine distance.}
\label{fig1}
\end{figure}

The attention mechanism is a procedure of assigning weights to the inputs which of late, have been modelled as a Markov Decision Process (MDP). However, in an NMT setup, attention mechanism can also be viewed as an election voting mechanism. An election can be represented as a pair $E = (C, V)$, where $C = (c_{1}, . . . , c_{m})$ is a set of candidates and $V = (v_{1}, . . . , v_{n})$ is a set of voters. Each voter provides a ranking of the candidates from the most to the least preferable. The winners are denoted by the set $W = (w_{1}, . . . , w_{k})$ which is a subset of $C$ and are elected using various election voting methods like SNTV, $k$-Borda, Bloc voting etc \cite{elkind2017properties}. Each of the $n$ output words of the translated sentence ($V$) give votes (attention weights) to the $m$ input words (of the original sentence) which can be, in turn, viewed as the candidates $C$.
We are specifically using $k$-Borda election method as it makes it easy to discover the extent to which the attention mechanism affects the input words (i.e. the election winners $W$ where $|W| = k$) which is signified by an optimal value of $k$. Subsequently, our experiments reveal that the attention weights can be modelled to some specific distribution, under some reasonable restrictions.

As mentioned earlier, attention mechanism is also a Markov Decision Process, owing to the fact that the result for next decoder step depends solely on the current decoder state (other than the inputs which remain fixed) of the NMT model, and not the previous ones. Taking inspiration from \cite{wang2018neural}, we attempt to introduce reinforcement learning techniques, specifically a Q-Learning model to aid the election mechanism in substituting an attention network. A state in a Markov Decision Process is characterised by various word embeddings in the vector space of the corpus. Being able to provide the best action leads to rewards (which signifies attention). The results from the voting method act as priors to the Q model.

\section{Previous Works}
Attention-based NMT models came into prominence after the work of \cite{bahdanau2014neural} which attempted to show that taking into consideration the weighted encoder inputs along with the current decoder state (instead of just the current decoder) provides more context for translation, thus performing better. This also led to an improvement in cases which involved context from words which are spread far across the sentence. By and by this research paved the way for the development of complicated models solely based on attention mechanism such as \cite{luong2015effective}, \cite{britz2017massive} and \cite{vaswani2017attention}. 

Apart from the usual domains where election voting methods find their use, they are also used in various other fields which involve some sort of social choice modelling such as \cite{Bordacou0:online} and \cite{chakraborty2019equality}. The former involves its use as a feature selection method whereas the latter finds its application in determining the fairness of recommender systems. 

Q-Learning finds its use in a variety of domains and exploits partially observable MDPs. Attempts to model the attention mechanism as an MDP can be found in works such as \cite{wang2018neural}. In this work, we try to leverage these ideas and make use of a Q-Learning to model the attention mechanism. 

We summarise the contributions of the paper as follows:-\\
1) We try to model attention weights to a known distribution under some assumptions assisted by $k$-Borda method.\\
2) Using Q-Learning method, the distribution initially obtained is tweaked to match the ground truth. \\
3) We also show that this approach yields significantly lower inference times to unseen data.

\begin{figure*}[t]
\centering
\includegraphics[width=\linewidth]{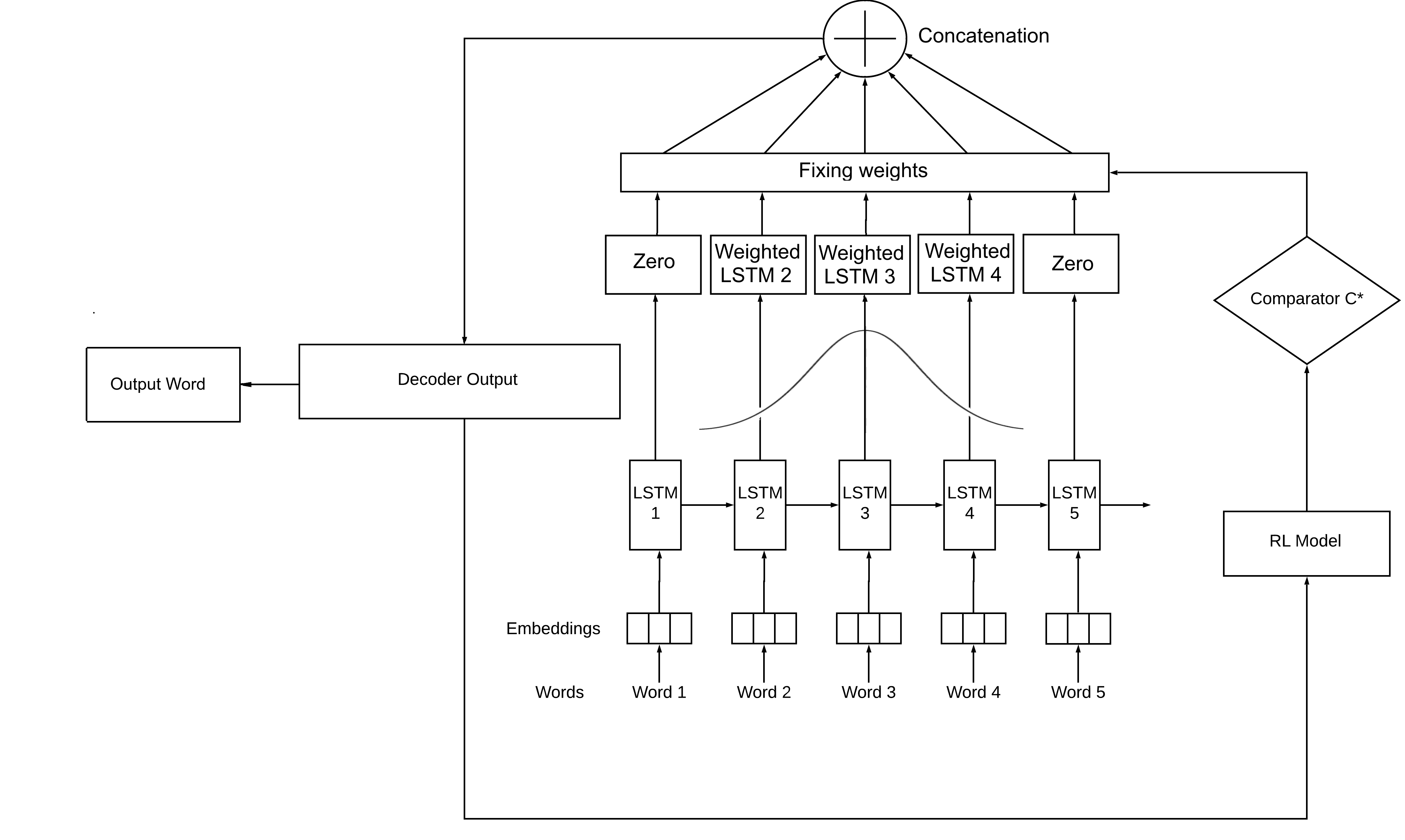} 
\caption{Proposed Architecture: The current situation represents the model decoding the third output word. Hence, the Gaussian curve appears over the third word in input as well. The RL output provides additional attention to words which the Gaussian curve might have missed.}
\label{fig1}
\end{figure*}

\begin{algorithm}
\caption{Inference algorithm}\label{alg:Inf}
\begin{algorithmic}[1]
\Procedure{Inf}{$string$
$wordsInSent, dict wordEmbeddings,$
$func Gaussian, func Q, model seq2seq$}
\State $embMat \gets []$
\State $translatedSentence = []$
\For{$ eachWord$ in $wordsInSent \do$}
\State $embWordEmb[eachWord]$
\State $push(embMatrix, emb)$
\EndFor
\State $out$, $hidden \gets seq2seq.encLSTM(embMatrix)$
\State $decoderOutput \gets out$
\For {$j = 0$ to $len(wordsInSent) -1 \do$}
\State $hiddenNow \gets Gaussian(hidden, j, len(wordsInSent)$
\State $listOfSimilarWords \gets Q(decoderInput)$
\State $i \gets 0$
\For {$eachWord$ in $wordsInSent \do$}
\If {$i == j$}
\State $continue$
\EndIf
\If {$eachWord$ is present in $listOfSimilarWords$}
\State $hiddenNow[i] \gets hiddenNow[i] + d * hidden[i]$ \Comment{$d$ was set as 0.25}
\EndIf
\State $i \gets i + 1$
\EndFor
\State $outWord \gets nearestWord(seq2seq.decLSTM(hiddenNow))$
\State $decoderOutput \gets wordEmbeddings[outWord]$
\State $push(translatedSentence, outWord)$
\EndFor
\State \textbf{return} $translatedSentence$
\EndProcedure
\end{algorithmic}
\end{algorithm}

\section{Proposed Methodology}

For the purpose of a proof of concept, we limit our experimentation to sentences with the same number of input and output words. Also, we try to model only Bahadanu Attention \cite{bahdanau2014neural} 
Network. Our approach involves two main steps. We describe them in detail below:

\subsection{Social Choice Theory modelling}
We used $k$-Borda count voting method to find out the extent among the input words (candidates) to which the attention weights are present for every output word (voters).  For example, in the translation ``esta es mi vida" to ``this is my life" the output word $my$ has attention weights from $es$, $mi$ and $vida$. In this case, we notice that for a four-word sentence, the $n^{th}$ output word has attention from $(n - 1)^{th}$, $n^{th}$ and $(n + 1)^{th}$ input words, i.e. $k = \lfloor log_{2}(p)\rfloor = 2$ where $p$ is the no of words in the sentence. Here, we are denoting the distance from the $n^{th}$ word on either of its sides up to which attention weights are significant as $k$. We tried to fit a Gaussian distribution to the weights, and it resembled in most of the cases. If otherwise, they are fine-tuned in the next step, which involves an MDP modelling of the same.


\begin{table*}[htbp]
\caption{Inference times in ms}
\centering
\scriptsize
\begin{tabular}{|c|c|c|c|}
\hline
{\textbf{Length of sentence}} & {\textbf{Attention}} & {\textbf{Gaussian Mask only}} & {\textbf{Gaussian Mask + RL model}} \\ \hline
4-7 & 1.46& 1.13 & 1.33  \\ \hline
     8-11 & 2.32 & 1.96 & 2.12 \\ \hline
12-15 & 2.86 & 2.43 & 2.61  \\ \hline

\hline
\end{tabular}
\label{}
\end{table*}

\subsection{Markov Decision Process modelling}
A Markov Decision Process is given by ($S$, $A$, $P$, $R$, $H$). In our model, $S$, which represents the set of states, is given by the entire input and output corpus, i.e. each state is a word embedding in its vector space. $A$ is the binary set of actions, whose presence denotes a movement of a hypothetical agent from one state to the other such that the corresponding words have a high cosine distance. Also, the agent moves only to a state which belongs to the input corpus. The action space contains two discrete values $\{0,1\}$ where $0$ signifies no action taken and $1$ signifies the agent hopping to the input word state yielding the maximum reward. The hypothetical agent is initially present at the current decoder's output word of the sequence to sequence model. Every time a word is decoded, the reward ($R$) is obtained as the cosine distance between the next decoder output and the predicted output. The agent is allowed to move to words belonging to the input corpus (Fig 1), which are close to each other such that the cosine distance between them is above a certain threshold. Words belonging to the input corpus are noted until the agent decides to take action $0$ (or till some finite steps) are obtained. Their presence is checked in the input set of words in the current translation, and when confirmed, a part of the attention weights is offered to such words. This step helped in fine-tuning the results from the election method, assisting in capturing words whose context might lie far outside the Gaussian mask. The horizon $H$ is finite. The Q value at time step $t$ is learnt as:
\newline
$Q(S_{t}, A_{t}) \Leftarrow Q(S_{t}, A_{t}) + \alpha[cosine$\textunderscore$distance(S_{t+1}, S_{t}) + max(Q(S_{t+i}, 1), Q(S_{t+i}, 0))  - Q(S_{t}, A_{t})]$\\

where $\alpha$ is the learning rate and $cosine$\textunderscore$distance$ calculates the cosine distance between embeddings of the given states.

\section{Experiments}
\subsection{Dataset Description}
We are using a language dataset provided by \cite{anki}, specifically the English-Spanish dataset. The corpus consists of 40,523 pairs of English and Spanish sentences after filtering the pairs having the same no. of words in both the languages.

We use a pretrained multilingual word embedding \cite{conneau2017word}
for this purpose. 

\subsection{Training}
We first train a Bahadanu Attention model on the mentioned corpus and embeddings. We perform max normalisation on the attention weights during training. The network was trained using Colab TPUs. The threshold for cosine distance was set to $0.79$.

\subsection{Validation}
After obtaining the attention weights on training, we use the validation set to initialise our model priors which consist of the value of $k$ and training the Q-model. We accumulate the Borda scores for the $(n - i)^{th}$ input words with $i$ $\epsilon $ $(0, \lfloor log_{s}(p))\rfloor$, where $p$ is the no of words in the sequence, and $s$ is varied between $2$ and $4$. This is followed by training the Q model. The network comprises three hidden fully connected layers with $128$, $256$ and $128$ neurons each. 

\subsection{Inference}
During inference, we use our proposed model with standard normal tweaked by the distortions caused using the learnt Q-learning model in place of the attention layer. We find the inference times are lower in the former case with negligible difference in BLEU scores.

\section{Results}
\subsection{Satisfaction and BLEU scores}
We choose a value of $k = \lfloor log_{2}(p)\rfloor$ for the Borda count metric as, beyond that, the increase was not too pronounced or involved too many words, as shown in Table 2. The BLEU scores were computed using equal n-gram weights. The results of the various models have been presented in Table 3. 
\subsection{Inference times}
The inference times are of the various models are mentioned in Table 1.

\begin{table}[htbp]
\caption{Satisfaction scores}
\centering
\scriptsize
\begin{tabular}{|c|c|c|c|}
\hline
{\textbf{Value of \textit{k}}} & {\textbf{Average Satisfaction}} \\ \hline
$\lfloor log_{4}(p) \rfloor$ & $55.01$  \\ \hline
$\lfloor log_{3}(p) \rfloor$ & $90.17$ \\ \hline
$\lfloor log_{2}(p) \rfloor$ & $93.22$  \\ \hline
$\lfloor p/2 \rfloor$ & $96.24$ \\ \hline

\hline
\end{tabular}
\label{}
\end{table}

\begin{table}[htbp]
\caption{BLEU scores}
\centering
\scriptsize
\begin{tabular}{|c|c|c|c|}
\hline
{\textbf{Model}} & {\textbf{        BLEU}} \\ \hline
Attention Layer (existing) & \textbf{24.22}  \\ \hline
     Gaussian Mask only & 23.02 \\ \hline
Gaussian Mask + RL model (proposed model) & \textbf{23.65}  \\ \hline

\hline
\end{tabular}
\label{}
\end{table}

\section{Conclusion and Future Work}
In this paper, we have experimented with an election mechanism and have depicted that a proper voting technique can mimic attention mechanism. In order to improve the results further, we modelled an MDP using Q-Learning. The results suggest that the proposed model performed only slightly below par in comparison to current attention models under restrictive assumptions. However, our model achieved lower inference times which is a clear indication of lower complexity and leading to a direction which could possibly lead to widespread adoption of attention mechanism. This acts as a proof of concept, yet, as a part of the future work, we plan to alleviate some of these restrictions by coming up with a better election method or remodelling the state space.
\newpage

\bibliographystyle{acl_natbib}
\bibliography{Bibliography}

\end{document}